\pdfoutput=1

\documentclass[11pt]{article}

\usepackage[final]{ACL2023}

\usepackage{times}
\usepackage{latexsym}
\usepackage{tikz}
\usepackage{adjustbox}
\usetikzlibrary{calc}
\usepackage{multirow}
\usepackage{multicol}
\usepackage[T1]{fontenc}

\usepackage[utf8]{inputenc}
\usepackage{amsmath}
\usepackage{amssymb}
\usepackage{breqn}
\usepackage{amsfonts}
\usepackage{algorithm}
\usepackage{algpseudocode}

\usepackage{microtype}
\usepackage{booktabs} %
\usepackage{xspace}
\usepackage[capitalise,noabbrev]{cleveref}
\usepackage{hyperref}
\usepackage{subcaption}

\newcommand{\sys}{PuMer\xspace} %

\title{\sys: Pruning and Merging Tokens for Efficient Vision Language Models}

\author{Qingqing Cao \\ 
\And
    Bhargavi Paranjape \\
\texttt{\{qicao,bparan,hannaneh\}@cs.washington.edu} \\
  University of Washington \\
\And 
   Hannaneh Hajishirzi \\
  }

\begin{document}
\maketitle
\begin{abstract}
Large-scale vision language (VL) models use Transformers to perform cross-modal interactions between the input text and image. These cross-modal interactions are computationally expensive and memory-intensive due to the quadratic complexity of processing the input image and text. We present \sys\footnote{Pronounced as ``puma''}: a token reduction framework that uses text-informed \textbf{P}r\textbf{u}ning and modality-aware \textbf{Mer}ging  strategies to progressively reduce the tokens of input image and text, improving model inference speed and reducing memory footprint. \sys learns to keep salient image tokens related to the input text and merges similar textual and visual tokens by adding lightweight token reducer modules at several cross-modal layers in the VL model. %
Training \sys is mostly the same as finetuning the original VL model but faster. 
Our evaluation for two vision language models on four downstream VL tasks shows \sys increases inference throughput by up to 2x and reduces memory footprint by over 50\% while incurring less than a 1\% accuracy drop. \footnote{Code is available at \url{https://github.com/csarron/PuMer}.}
\end{abstract}
\section{Introduction}
\label{sec:intro}
Large-scale vision language models~\cite{douEmpiricalStudyTraining2021,wangOFAUnifyingArchitectures2022,zengMultiGrainedVisionLanguage2021,kimViLTVisionandLanguageTransformer2021,wangSimVLMSimpleVisual2021,zhangVinVLRevisitingVisual2021} have shown substantial progress on many vision language tasks such as visual question answering, natural language visual reasoning, and visual entailment. 
However, state-of-the-art language and vision models are memory intensive and computationally expensive because they use multi-layer self-attention between many language and vision input tokens (small image patches) with quadratic complexity. This inefficiency 
limits high-throughput cloud deployments and makes it infeasible to run on resource-constrained devices. 

\begin{figure}[t!]
	\begin{center}
		\includegraphics[width=0.48\textwidth]{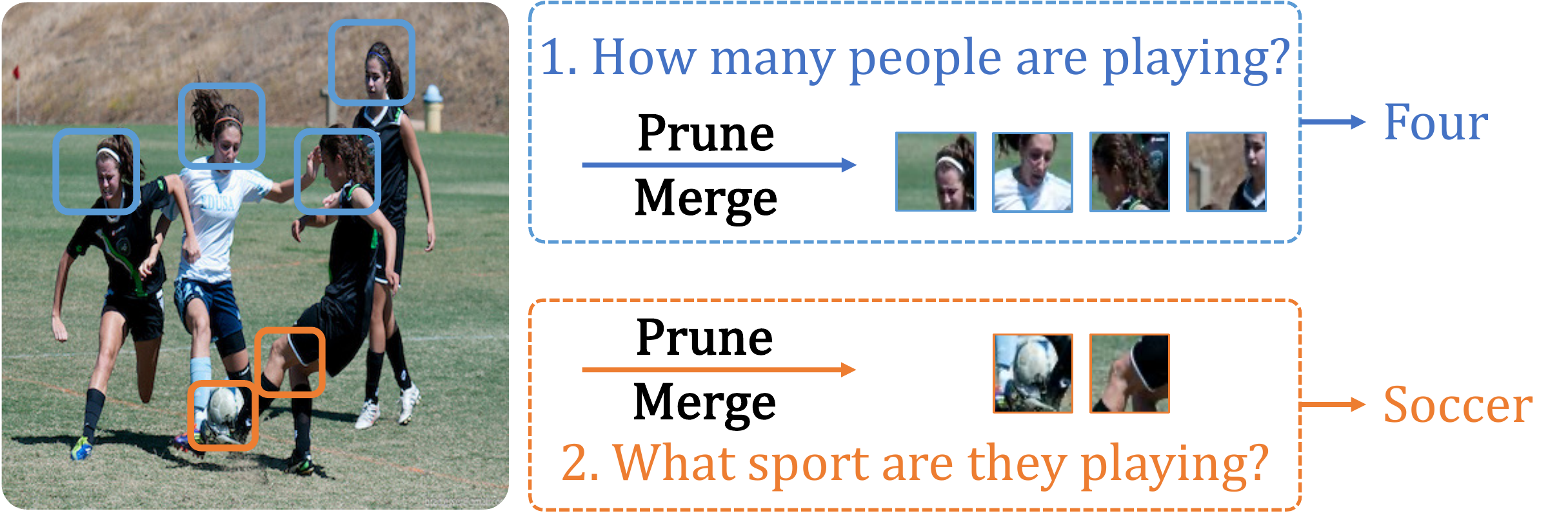}
		\caption{\sys applies token reduction to VL models via pruning and merging. \sys makes VL models run faster by text-informed image pruning to remove text-irrelevant image tokens and modality-aware merging to compress similar input tokens. }
		\label{fig:eg}
	\end{center}
\end{figure}
The key source of inefficiency in deep VL models is that these models need to process the entire input image and text tokens over all the layers. Our intuition is that  the input image contains redundant information %
and only parts of the image (\emph{salient} regions, referred by the text) are required and related to the end task. %
For example, in \cref{fig:eg},  most of the image content (the four persons, field) is not needed except for the bottom-center soccer region to answer the visual question ``What sport are they playing?''. This paper advocates using the correlations between image and text modalities to reduce  tokens for VL problems. 

In the vision-only or text-only domains, researchers have shown that reducing image  or text tokens can improve the model computational complexity through \emph{pruning}  ~\cite{liangEViTExpeditingVision2021,raoDynamicViTEfficientVision2021,yinAViTAdaptiveTokens2022,marinTokenPoolingVision2021,goyalPoWERBERTAcceleratingBERT2020a} that learns to remove non-salient image or text tokens for a given task; or \emph{merging} ~\cite{bolyaTokenMergingYour2022,xuGroupViTSemanticSegmentation2022,ryooTokenLearnerWhatCan2021} that groups semantically similar tokens. Using either reduction method in isolation is not sufficient for a VL problem setting since \textit{i)} salient image tokens are different given different text inputs, \textit{ii)} pruning alone causes big information loss, hurting the performance, \textit{iii)} merging tokens irrespective of their modality confuses the VL models since text and image token representations cannot be perfectly aligned to the same semantic space. In this paper, we design a lightweight and effective framework that integrates these token reduction strategies into VL models.

We introduce \textbf{\sys}, a token reduction framework that consists of \textbf{P}r\textbf{u}ning-and-\textbf{Mer}ging operations to gradually reduce image tokens that are not related to text and merge image and text tokens respective to their modality.  In particular, we design \emph{(i) text-informed image token pruning} to remove image tokens that are irrelevant to text and are unimportant to the VL task predictions (removing tokens that describe persons and field for the second question in the \cref{fig:eg} example); \emph{(ii) modality-aware token merging} to merge semantically redundant tokens for text and image tokens modality independently (combining the image tokens describing each person for the first question in \cref{fig:eg}). We keep the remaining tokens that are neither pruned nor merged. 
At the core of \sys is a set of lightweight non-parametric token reducers that decide which image tokens are pruned and merged as the VL model forward computation proceeds. To reduce abrupt image information loss and improve computational efficiency, we scatter the token reducers at different cross-modal layers in the VL model and reduce the tokens in a cascaded fashion. Fewer tokens are pruned and merged in earlier layers.

\sys is easy to train since the token reducers contain no parameters and add little overhead. The training procedure is  almost the same as finetuning the original VL models, except that we add a knowledge distillation loss that further reduces the accuracy gap compared to finetuned models. Though we focus on inference efficiency, \sys makes VL models run faster for both training and inference because text and image tokens are reduced in the forward computation.

We evaluate \sys over two recent VL models ViLT~\cite{kimViLTVisionandLanguageTransformer2021} %
 and METER~\cite{douEmpiricalStudyTraining2021} across five vision language tasks: text-image retrieval tasks (including image-to-text and text-to-image retrieval)~\cite{plummerFlickr30kEntitiesCollecting2015}, visual question answering (VQAv2;~\citealt{goyalMakingVQAMatter2017}), natural language visual reasoning (NLVR2;~\citealt{suhrCorpusReasoningNatural2019}), and visual entailment (SNLI-VE;~\citealt{xieVisualEntailmentTask2019}). Compared to baselines, \sys improves the model inference throughput by \textbf{1.7x}$\sim$\textbf{2.1x} and reduces memory footprint by \textbf{38\%}$\sim$\textbf{50\%} with minimal (less than 1\%) accuracy loss. 
 Our analysis validates that both text-informed image pruning and modality-aware token merging contribute to the token reduction effectiveness of \sys.

\section{Related work}
\label{sec:related}

\paragraph{Token Reduction in NLP and Vision.} Prior work in data pruning ~\cite{raoDynamicViTEfficientVision2021,yinAViTAdaptiveTokens2022,liangEViTExpeditingVision2021,goyalPoWERBERTAcceleratingBERT2020a} focus on single-modality models by either pruning input text or image alone. DynamicViT ~\cite{raoDynamicViTEfficientVision2021} and A-ViT~\cite{yinAViTAdaptiveTokens2022} both progressively remove the uninformative content and keep salient regions in the input image. This type of pruning does not apply to language and vision tasks where the salient regions depend on the input text. Our work shows different input texts lead to pruning different image regions even for the same input image. 
PoWER-BERT~\cite{goyalPoWERBERTAcceleratingBERT2020a} speeds up the inference of text-based Transformers like BERT~\cite{devlinBERTPretrainingDeep2019b} by removing the input text tokens, which are not the main computation bottlenecks for most vision and language tasks. 

Another line of work seeks to reduce input tokens by merging tokens. SPViT~\cite{kongSPViTEnablingFaster2022} and EViT~\cite{liangEViTExpeditingVision2021} select uninformative image tokens and combine them into one token. And EViT also requires expensive pretraining. GroupViT~\cite{xuGroupViTSemanticSegmentation2022} combines image tokens via cross-attention to find similar objects for semantic segmentation. Recently, ToMe~\cite{bolyaTokenMergingYour2022}, TokenLearner~\cite{ryooTokenLearnerWhatCan2021} and TokenPooling~\cite{marinTokenPoolingVision2021} combine tokens without pruning and achieves better speedup versus accuracy trade-offs. 

Our method is inspired by token pruning and merging works but integrates them into a token reduction framework suitable for VL models. Our key difference is to leverage the relationships between textual and visual tokens to remove and combine tokens. Our experiments (\cref{sec:eval}) show improvements over these lines of work.

\paragraph{Efficient Vision Language Models.} Many techniques have focused on model pruning~\cite{lagunasBlockPruningFaster2021,yuUnifiedPruningFramework2021,yu2022width,TPruneEfficientTransformer}, dynamic computation by early exiting~\cite{xinEarlyExitingBERT2020,NEURIPS2020_d4dd111a,schwartzRightToolJob2020,liuFastBERTSelfdistillingBERT2020a,caoMobiVQAEfficientOnDevice2022} or designing small and efficient VL models~\cite{fangCompressingVisualLinguisticModel2021,wangMiniVLMSmallerFaster2020}. Combining these orthogonal optimizations with our token reduction method could further accelerate the inference in VL models. %

\section{Background and Overview}
\label{sec:background}

\begin{figure}[h!]
	\begin{center}
		\includegraphics[width=0.4\textwidth]{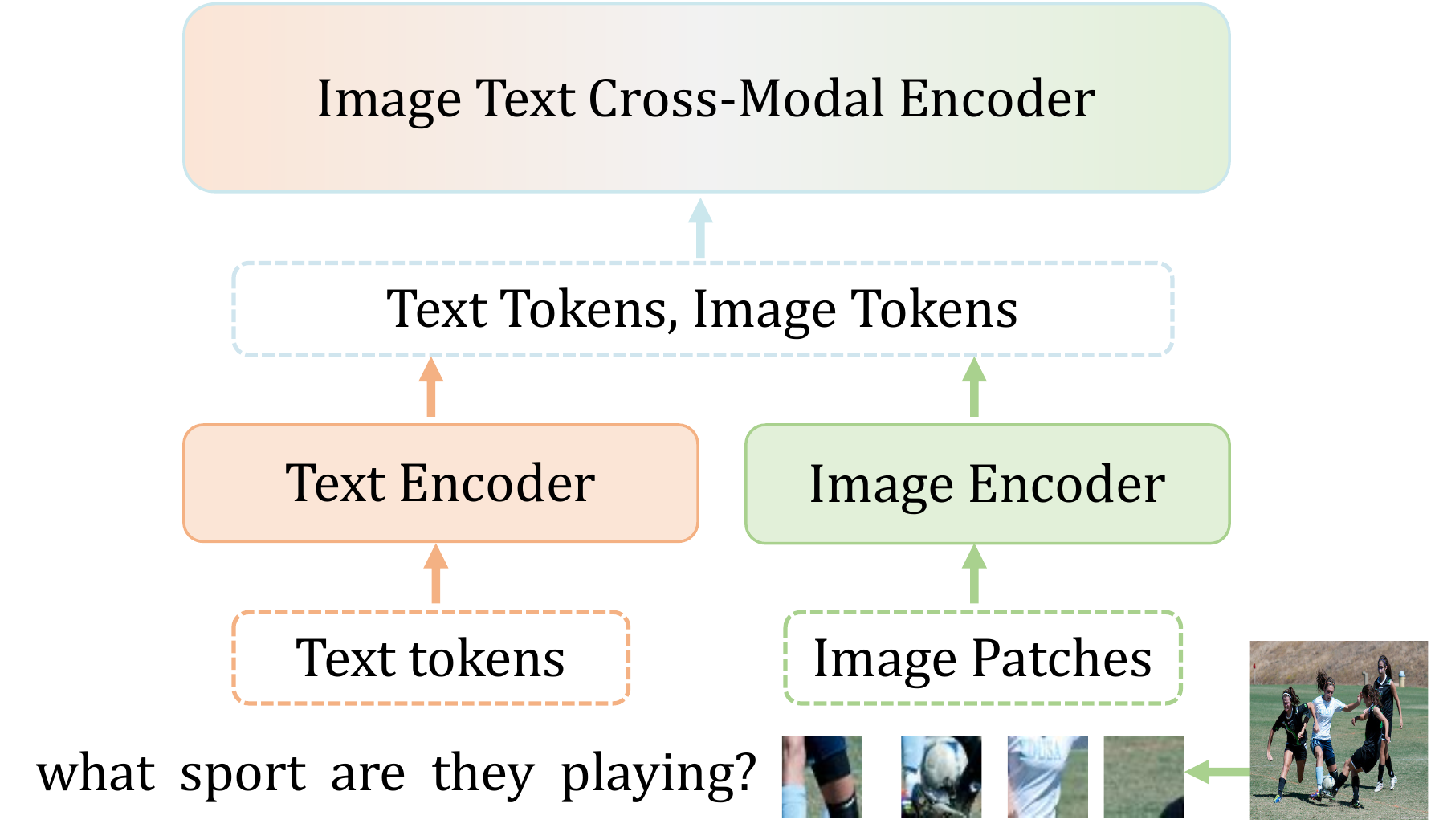}
		\caption{General architecture of vision language models. The input image is projected into many small image patches (``tokens'') that are processed by the image encoder. The cross-modal  attention between text and image tokens has quadratic time complexity,
		 which is computationally expensive. Both ViLT and METER models follow this pattern.}
		\label{fig:vl-arch}
	\end{center}
\end{figure}

\paragraph{Vision Language Models.} \cref{fig:vl-arch} shows the backbone of a VL model consisting of a text encoder, an image encoder, and a cross-modal encoder. The input sentence (e.g. a question or a statement) is first tokenized as text tokens and fed to the text encoder to create contextualized text representations. Similarly, the input image is projected into many small image patches, referred to as ``image tokens'',  that are further contextualized by the image encoder. Finally, the cross-modal encoder takes the concatenated text and image tokens and fuses information between image and text modalities via Transformer-style~\cite{vaswani2017attention} cross-attention interactions. 

 For many VL tasks, the number of tokens of the input image is an order of magnitude more than that of the input text --- a visual question can have at most a dozen tokens but the associated  image consists of a hundred image tokens. For example, for an image with a resolution of 384x384 and a patch size of 16, the number of tokens is $(384/16)^2=576$.

\paragraph{Token Reduction for Efficiency.} In this paper, we focus on {\it reducing} image tokens to improve computational efficiency of the model through {\it pruning} and {\it merging}. However,  naively removing a large percentage of the image tokens inside the cross-modal layers may cause abrupt image information loss, as the VL model is trained to build representations of the full image for the downstream task. 
For example, if the soccer region in \cref{fig:eg} gets pruned, the VL model is unlikely to output the answer ``soccer'' for the question ``what sport are they playing?''. On the other hand, simply merging image tokens without text guidance can lead to suboptimal performance. For example, merging the image regions of the background field and soccer in \cref{fig:eg} does not contribute to answering the visual question ``how many people are playing?''.

The next section describes our text-informed token reduction approach. The basic building blocks of \sys are lightweight non-parametric {\it token reducers} that reduce image and text tokens in a cascaded manner to mitigate the information loss and improve the computational efficiency of a VL model.

 \begin{figure*}[ht!]
	\begin{center}
		\includegraphics[width=\linewidth]{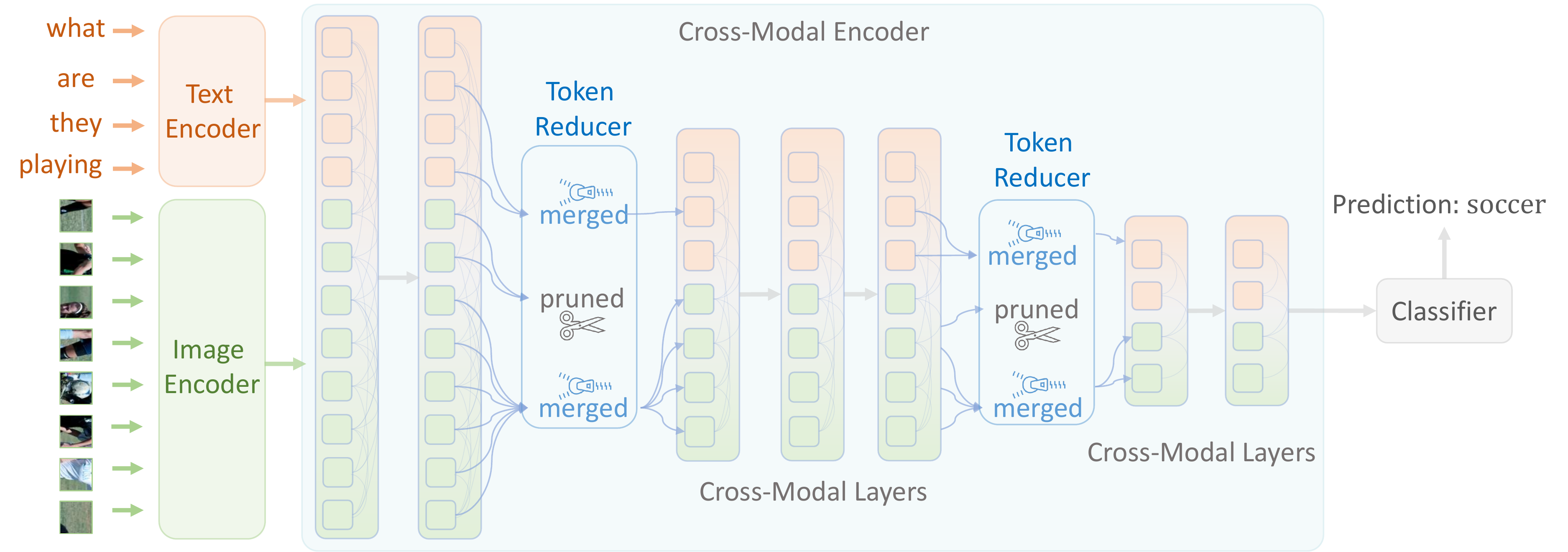}
		\caption{ \sys applies token reducers in the cross-modal layers of a VL model. Each token reducer is non-parametric and uses text-informed pruning and modality-aware merging to reduce image and text tokens.}
		\label{fig:tip}
	\end{center}
\end{figure*}

\section{\sys: Text-Informed Token Reduction Framework}
\label{sec:method}

\label{sec:design}
Given a VL cross-modal encoder, 
 \sys progressively reduces image tokens going through the cross-modal encoder  (depicted in \cref{fig:tip}). \sys uses lightweight token reducers with no learnable parameters, adding them in different layers of the cross-modal encoder to predict which image tokens are removed or merged. 

\begin{algorithm*}[ht!]
\caption{Token Reduction via Text-Informed Image Pruning and Modality-Aware Merging}
\label{algo:tip}
\textbf{Input:} text token vectors $\mathbf{T}$, text-to-image cross
 attention scores $\mathbf{A}$,  image token vectors $\mathbf{V}$, \newline prune ratio $k$, image merge ratio $r$, text merge ratio $t$ \\
\textbf{Output:} merged text token vectors $\mathbf{T}_m$,  pruned and merged image token vectors $\mathbf{V}_m$

\begin{algorithmic}[1]
\State for image tokens $\mathbf{V}$, compute text-saliency scores $\mathbf{s}$ using Eq\ref{eq:text-saliency}; \Comment{text-informed image pruning}
\State obtain indices $idx$ of top-$k^{\prime}$ items in score $\mathbf{s}$, $k^{\prime}=(1-k)|\mathbf{V}|$; \Comment{$k^{\prime}$ is the \# of kept image tokens}
\State select $k^{\prime}$ image tokens by the top-$k^{\prime}$ indices, $\mathbf{V}_p=\mathbf{V}[idx]$;
\State merge text tokens $\mathbf{T}$ by bipartite soft matching into $\mathbf{T}_m=\text{bipartite_merge}(\mathbf{T}, t)$; \newline merge image tokens $\mathbf{V}_p$ into $\mathbf{V}_m=\text{bipartite_merge}(\mathbf{V}_p, r)$
\Comment{modality-aware merging}
\Procedure{bipartite_merge}{input tokens: $\mathbf{X}$, merge ratio: $r$}
  \State divide tokens $\mathbf{X}$ into two sets of tokens $\mathbf{O}$ and $\mathbf{E}$ based on even and odd order
  \State for each token $\mathbf{O}_a$ in $\mathbf{O}$, compute its top-1 similar token $\mathbf{E}_b$ in $\mathbf{E}$, save the indices $a$ and $b$ into\newline
  \hspace*{\algorithmicindent}a token edge (an edge between $\mathbf{O}_a$ and $\mathbf{E}_b$), save all token edges as $\mathbf{P}$ and corresponding \newline
  \hspace*{\algorithmicindent}top-1 similarity scores $\mathbf{S}_p$ 
  \Comment{this can be implemented as a fast parallel operation}
  \State $r^{\prime}=r|\mathbf{X}|$, obtain indices $ind$ of top-$r^{\prime}$ items in $\mathbf{S}_p$, select top-$r^{\prime}$ edges: $\mathbf{P}_r = \mathbf{P}[ind]$
 \State for each token edge ($a$, $b$) in $\mathbf{P}_r$, collect tokens from $\mathbf{O}$ and $\mathbf{E}$, merge tokens in $\mathbf{O}$ and $\mathbf{E}$ that \newline
  \hspace*{\algorithmicindent}are connected via edges (sharing the same token as a vertex node) into $\mathbf{OE}$ by computing \newline
  \hspace*{\algorithmicindent}the average of each token vectors, gather $\mathbf{O}_{rest}$ and $\mathbf{E}_{rest}$ from the rest (unmerged) indices. 
  \State output: merged tokens $\mathbf{X}_m = gather(\mathbf{OE}, \mathbf{O}_{rest}, \mathbf{E}_{rest})$
\EndProcedure
\end{algorithmic}
\end{algorithm*}

\paragraph{Token Reducers.} %
For an $n$-layer cross-modal encoder, after the first $f$ ($f<n$) layers, a token reducer first removes $k$\% of the image tokens at any layer $\ell$ between $f$ and $n$ guided by the text information. The tokens removed in layer $\ell$ are not used in subsequent layers. Then the token reducer merges $r$\% and $t$\% of the image and text tokens respectively in layer $\ell$.  We scatter the token reducers across the cross-modal layers to achieve a better accuracy and efficiency trade-off. Intuitively, reducing at early layers in the cross-modal encoder will have higher inference efficiency but may have bigger performance loss and vice versa. We study this trade-off in more detail in \cref{sec:ablation}.

The token reduction algorithm is described in \cref{algo:tip}. Each token reducer consists of two sequential non-parametric modules: first, a \emph{text-informed pruner}~(\textbf{TIP}) prunes image tokens that are not related to the accompanying text  (\cref{sec:tip}); second, a \emph{modality-aware merger}~(\textbf{MAM})  reduces tokens by merging similar tokens within the image or text modality (\cref{sec:mam}).
These two steps reduce the image and text tokens to benefit the computational efficiency, while not losing the accuracy. 
Note that if we only apply text-informed pruning to the images without merging, to achieve similar efficiency gains, we need to set a larger pruning ratio which will hurt task performance due to substantial information loss. Instead of dropping such information, modality-aware merging helps alleviate information loss by compressing semantically similar content into fewer tokens while still providing efficiency benefits.

\subsection{Text-Informed Image Pruning}\label{sec:tip}

The first step is to prune image tokens according to their relevance to the text. The intuition is that only some parts of the image are important for the end language-vision task, hence removing the text-irrelevant parts will not hurt the performance, while it improves the computational efficiency.  %
Unlike previous works~\cite{raoDynamicViTEfficientVision2021} that use extra learnable parameters to predict which image tokens to prune, we take a different but faster approach without using any parameters. The key idea is to use the text-to-image cross-attention scores\footnote{VL models use cross-attention to perform information fusion between different modalities.} that are already available in the VL model to compute how important each image token is to the text. We keep important image tokens and prune the rest. Since this text-informed pruning also removes image tokens during training, it trains faster\footnote{We observe 15\%$\sim$20\% faster training speed in practice.} than parameter-based pruning approaches like~\citet{raoDynamicViTEfficientVision2021}. 

For each cross-modal layer $\ell$ where the token reducer is applied, we denote the input text token vectors as $\mathbf{T}$, image token vectors as $\mathbf{V}$, and text-to-image cross-attention scores as $\mathbf{A}$ (computed in the cross-attention layer that already exists in a VL model). We first compute the text-saliency scores $\mathbf{s}$ for every image token:
\begin{equation} \label{eq:text-saliency}
s_v=\frac{1}{|T|}\sum^{t=1}_{|T|}\sum^{h=1}_{H}\mathbf{A}^{h}_{tv},
\end{equation}
where $|T|$ is the number of text tokens, $H$ the number of attention heads, $t$ and $v$ are the indices of text and image tokens. This text-saliency score for the image token is text-informed because each value is summed over all text tokens, and an image token with a bigger text-saliency score means it's attended more by the text and hence is more text-relevant. Next, we keep top-$k^{\prime}$ image tokens\footnote{$k^{\prime}=(1-k)|\mathbf{V}|$ is the number of kept tokens}  $\mathbf{V}_p$ according to their text-saliency score and discard the remaining image tokens.

\subsection{Modality-Aware Merging}\label{sec:mam}
Once text-irrelevant image tokens are pruned, the remaining image tokens contain more text-salient information but they might still be redundant. 
For example, multiple image tokens describe the same person  in the \cref{fig:eg} image and their representations might be similar (their vector distances are close). For the text modality, the token redundancy still exists due to the self-attention contextualization which progressively creates similar information~\cite{goyalPoWERBERTAcceleratingBERT2020a}. In practice, text tokens are padded to max length for efficient training and inference, these padding tokens also contribute to redundancy.

In this section, we describe our modality-aware merging approach to eliminate such redundancy. In particular, our method merges semantically similar image tokens $\mathbf{V}_p$ into a single image token and similar text tokens $\mathbf{T}$ into a single text token to further reduce the number of tokens. We specifically merge tokens within each modality, i.e., image tokens are merged with similar image tokens, and text tokens are merged with similar text tokens. %

To implement modality-aware merging, we need to identify similar tokens and combine their information in a lightweight way. Existing methods such as k-means clustering~\cite{marinTokenPoolingVision2021}, pooling~\cite{pietruszkaSparsifyingTransformerModels2020,nawrotEfficientTransformersDynamic2022}, grouping~\cite{xuGroupViTSemanticSegmentation2022} or learning-based~\cite{ryooTokenLearnerWhatCan2021} cause non-negligible overhead and slow down the VL model computation, instead, we use the bipartite soft matching algorithm~\cite{bolyaTokenMergingYour2022} to find similar tokens and combine them in parallel.

\begin{figure}[h!]
	\begin{center}
		\includegraphics[width=0.45\textwidth]{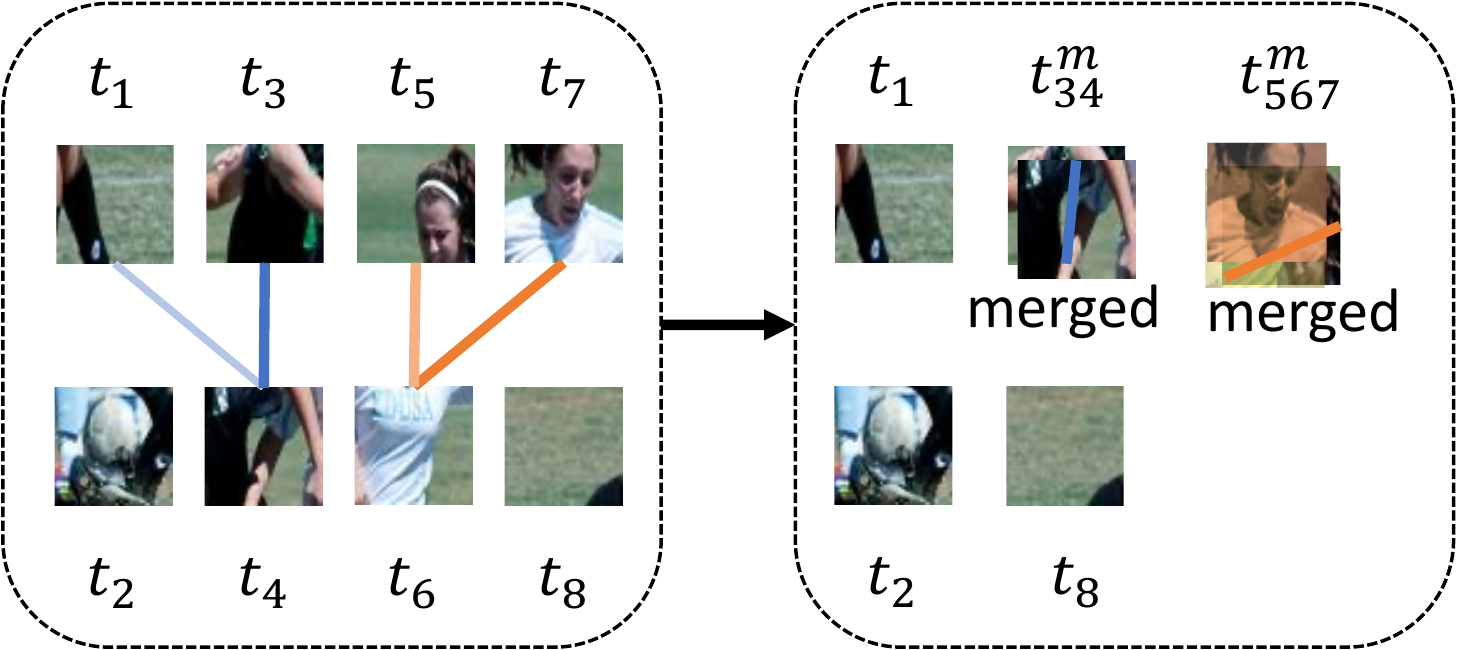}
		\caption{Illustration of merging by bipartite matching. In this example, there are 8 tokens, $\mathbf{E}$ consists of token $t_1$, $t_3$, $t_5$ and $t_7$, $\mathbf{O}$ has $t_2, t_4, t_6, t_8$. Assume for $t_1, t_3, t_5, t_7$ in $\mathbf{E}$, the most similar tokens in $\mathbf{O}$ are $t_4, t_4, t_6, t_6$ respectively, and $t_3-t_4, t_7-t_6, t_5-t_6$ are the edges (darker and thicker lines mean larger similarity values) with top-$r^{\prime}$ ($r^{\prime}=3$) most similarity, then we merge $(t_3, t_4)$ into one token $t^m_{34}$, $(t_5, t_7, t_6)$ into one token $t^m_{567}$, and keep $t_1, t_2, t_8$, in this case, we reduce three (3/8=37.5\%) tokens.}
		\label{fig:bipartite}
	\end{center}
\end{figure}

Here, we explain the bipartite matching approach in more detail. Specifically, the inputs are a set of token vectors $\mathbf{X}$ (can be $\mathbf{V}_p$ or $\mathbf{T}$) and a merge ratio $r$, we form a bipartite graph by dividing the nodes (tokens) into two disjoint sets (say $\mathbf{E}$ and $\mathbf{O}$) of equal size based on their order (even or odd). Then, for each token in $\mathbf{O}$, we find its most similar token in $\mathbf{E}$, and draw an edge between the token pair (lines in the left figure in \cref{fig:bipartite}). We select the top-$r^{\prime}$ edges\footnote{$r^{\prime}=r|\mathbf{X}|$ is the number of merge tokens} %
based on the similarity and merge their corresponding (most similar) token in $\mathbf{E}$ and $\mathbf{O}$. \cref{fig:bipartite} shows an example of bipartite matching. Since the self-attention in a VL model layer already has computed keys and values for each token to measure similarity, following~\citet{bolyaTokenMergingYour2022}, %
we compute the similarity as the dot product $\mathbf{S}_p^{t_1t_2}=\mathbf{K}_{t_1} \mathbf{K}_{t_2}$ between the keys of each token vector $\mathbf{X}_i$. We keep the rest non-top-$r^{\prime}$ tokens in $\mathbf{O}_{rest}$ and unmerged tokens in $\mathbf{E}_{rest}$. We also describe this procedure in \cref{algo:tip}. 

\subsection{Training and Inference}
Token reducers in \sys contain no trainable parameters and can be incorporated into off-the-shelf VL models without changing model architectures for both training and inference. \sys is easy to train and follows the same setup as finetuning original VL models. To reduce the accuracy drop further, we add a knowledge distillation~\cite{hintonDistillingKnowledgeNeural2015a} loss. During training and inference, \sys has three configurable hyperparameters (keep ratio $k$, merge ratios $r$, and $t$ for image and text) to control the efficiency versus accuracy trade-offs.

{\noindent \bf Implementation Details.} We set the pruning and merging ratio in the range of 0.1 to 0.5 in 3 or 4 locations in cross-modal layers. The exact values are in~\cref{sec:impl}. In \cref{sec:ablation}, we study the design choices for different reduction ratios and reduction layer locations. More implementation and training details are in \cref{sec:impl}.

\section{Evaluation Setup}
\label{sec:eval}

\subsection{Backbone Vision-Language Models}

We evaluate \sys for two different VL models: ViLT~\cite{kimViLTVisionandLanguageTransformer2021} with 110 million parameters and a state-of-the-art VL model, 
METER~\cite{douEmpiricalStudyTraining2021} with 330 million parameters.
We denote \sys-ViLT and \sys-METER as \sys applied for ViLT and METER respectively. %

\paragraph{ViLT} is a recent efficient VL model that uses BERT~\cite{devlinBERTPretrainingDeep2019b} embeddings to encode text and a linear layer to project image patches. ViLT then concatenates the text and image tokens and uses a 12-layer Transformer encoder to perform the cross-modal fusion. ViLT is a relatively lightweight model and has 110 million parameters.

\paragraph{METER} is a state-of-the-art VL model that uses RoBERTa~\cite{liuRoBERTaRobustlyOptimized2019a} as the text encoder and CLIP~\cite{radfordLearningTransferableVisual2021a} as the image encoder, and 12 BERT-like cross-attention layers to fuse the text and image modalities. METER is a large model and has 330 million parameters. 

\subsection{Evaluation Tasks}
We evaluate the models on five vision-language language tasks: 

\paragraph{Image-Text Retrieval} contains two subtasks: image-to-text retrieval (IR) and text-to-image retrieval (TR). We finetune \sys and evaluate on the Flickr30K~\cite{plummerFlickr30kEntitiesCollecting2015}. 

\paragraph{Visual Question Answering (VQAv2)} dataset ~\cite{goyalMakingVQAMatter2017} contains over 1 million diverse open-ended questions about images both from the MSCOCO~\cite{linMicrosoftCOCOCommon2014} and real-world scenes. Answering these questions requires an understanding of vision, language, and commonsense knowledge. 

\paragraph{Visual Entailment (VE)}~\cite{xieVisualEntailmentTask2019} is a visual inference task that consists of 570K sentence image pairs constructed from the Stanford Natural Language Inference corpus~\cite{bowmanLargeAnnotatedCorpus2015} and Flickr30k~\cite{youngImageDescriptionsVisual2014}. The goal is to predict whether the image premise semantically entails the text. 

\paragraph{Natural Language for Visual Reasoning (NLVR2)} corpora ~\cite{suhrCorpusReasoningNatural2019} have over 100K examples of linguistically diverse English sentences written by humans and are grounded in pairs of visually complex images. The goal is to predict whether a sentence is true about two input images.

\begin{table*}[t]
    \centering
    \setlength\tabcolsep{7pt}
    \small
\begin{tabular}{@{}llcccc@{}}
\hline
Model        & Datasets     & Original Accuracy & \sys Accuracy & Throughput Increase & Memory Reduction \\ \hline
 \multirow{5}{*}{METER (SoTA)}             & Flickr30k TR & 94.7                     & 93.8 (-0.9)                         & 1.81x                & 38\%              \\
             & Flickr30k IR & 82.0                     & 81.2 (-0.8)                         & 1.81x               & 38\%              \\
             & VQAv2        & 77.5                     & 76.8 (-0.7)                  & 1.82x              & 38\%             \\
  & SNLI-VE           & 81.1                     & 80.3 (-0.8)                  & 2.07x               & 43\%             \\
             & NLVR2        & 82.7                     & 82.2 (-0.5)                  & 1.79x               & 38\%             \\ \hline
 \multirow{5}{*}{ViLT}             & Flickr30k TR & 78.2                     & 77.6 (-0.6)                         & 1.78x                & 46\%              \\
             & Flickr30k IR & 60.2                     & 59.6 (-0.7)                         & 1.78x                & 46\%              \\
             & VQAv2        & 69.5                     & 68.9 (-0.6)                  & 1.76x               & 45\%             \\
         & SNLI-VE           & 76.0                     & 75.6 (-0.4)                  & 2.01x               & 51\%             \\
             & NLVR2        & 75.5                     & 74.9 (-0.6)                  & 1.74x               & 45\%             \\ \hline
\end{tabular}
\caption{Performance and inference efficiency comparison between the original fine-tuned vs \sys fine-tuned models for the ViLT and METER over four downstream visual reasoning tasks. }
\label{table:main-results}
\end{table*}

\subsection{Baselines}
To compare the benefits of \sys, we additionally evaluate three baselines:

\noindent \textbf{DynamicViT}~\cite{raoDynamicViTEfficientVision2021} designs several prediction modules parameterized by MLPs to predict which image tokens to prune in vision transformers~\cite{dosovitskiyImageWorth16x162020a}. For a fair comparison, we use the original DynamicViT configurations (pruning layers and ratios) for the ViLT model. %

\noindent \textbf{ToMe}~\cite{bolyaTokenMergingYour2022} uses token merging to reduce the number of tokens in vision transformers. We configure ToMe to make sure similar speedup as \sys and compare their accuracy.

Note that both DynamicViT and ToMe are designed for vision Transformers and work for image modality, therefore they do not distinguish between the image and text tokens. On the contrary, \sys is a more general token reduction framework that uses text to guide the image pruning and makes merging modality aware.  

\noindent \textbf{Smaller Resolution} (SmRes): We downsample the input image to smaller resolutions and finetune the VL models. Using smaller input images directly reduces the computation of VL models.

\subsection{Evaluation Metrics}

\paragraph{Accuracy Metrics.} We measure \emph{VQA accuracy}~\cite{goyalMakingVQAMatter2017} for the VQAv2 dataset and \emph{accuracy} for both the VE and NLVR2 datasets. For text retrieval (TR) and image retrieval (IR) tasks, the accuracy refers to Top1-recall.
Unlike  previous works~\cite{kimViLTVisionandLanguageTransformer2021,douEmpiricalStudyTraining2021}, where their models are trained on the combined training and validation sets, our focus is not to obtain state-of-the-art results, so we train the two VL models on the training set and report the results on the test set. All the accuracy numbers are average values across 3 runs. 

\paragraph{Efficiency Metrics.} We measure the actual inference throughput (examples per second) of the VL models on the GPU hardware and compare them to the original finetuned models, and we report the \emph{throughput increase}. We also measure the peak memory consumed during the model inference phase and report \emph{memory reduction} ratio compared to the original finetuned models. These two runtime metrics reflect actual efficiency and are found to be more accurate to compare resource consumption instead of using the FLOPs complexity metric~\cite{grahamLeViTVisionTransformer2021}. For comparison purposes, we include the FLOPs comparison in the appendix \cref{sec:flops}.

For inference throughput measurements, we increase the batch size until the model gets out of GPU memory, and run the inference with the batch size that gives the biggest throughput for 30 seconds on a single GPU. For inference memory footprint, we use the same batch size for the original VL model and \sys version and report the peak memory difference. For ViLT models, we use GTX 1080 Ti GPU and start the batch size from 32 with a step of 8; for METER models, we use an A40 GPU and start the batch size from 16 with a step of 8. %

\section{Experimental Results}
\subsection{Main Results}

\paragraph{\sys is faster and remains accurate.} \cref{table:main-results} shows the main results comparing performance, inference speed, and memory reduction of \sys~ versus the original models.  Overall, we observe over \textbf{1.7x $\sim$ 2x speedup} in inference throughput and over \textbf{35\% $\sim$ 51\% reduction} in memory footprint for both ViLT and METER models on the VL tasks. Importantly, the task performance of \sys remains competitive compared to the original finetuned VL models with only $<$1\% drop in accuracy.

\begin{figure}[t]
	\begin{center}
		\includegraphics[width=\linewidth]{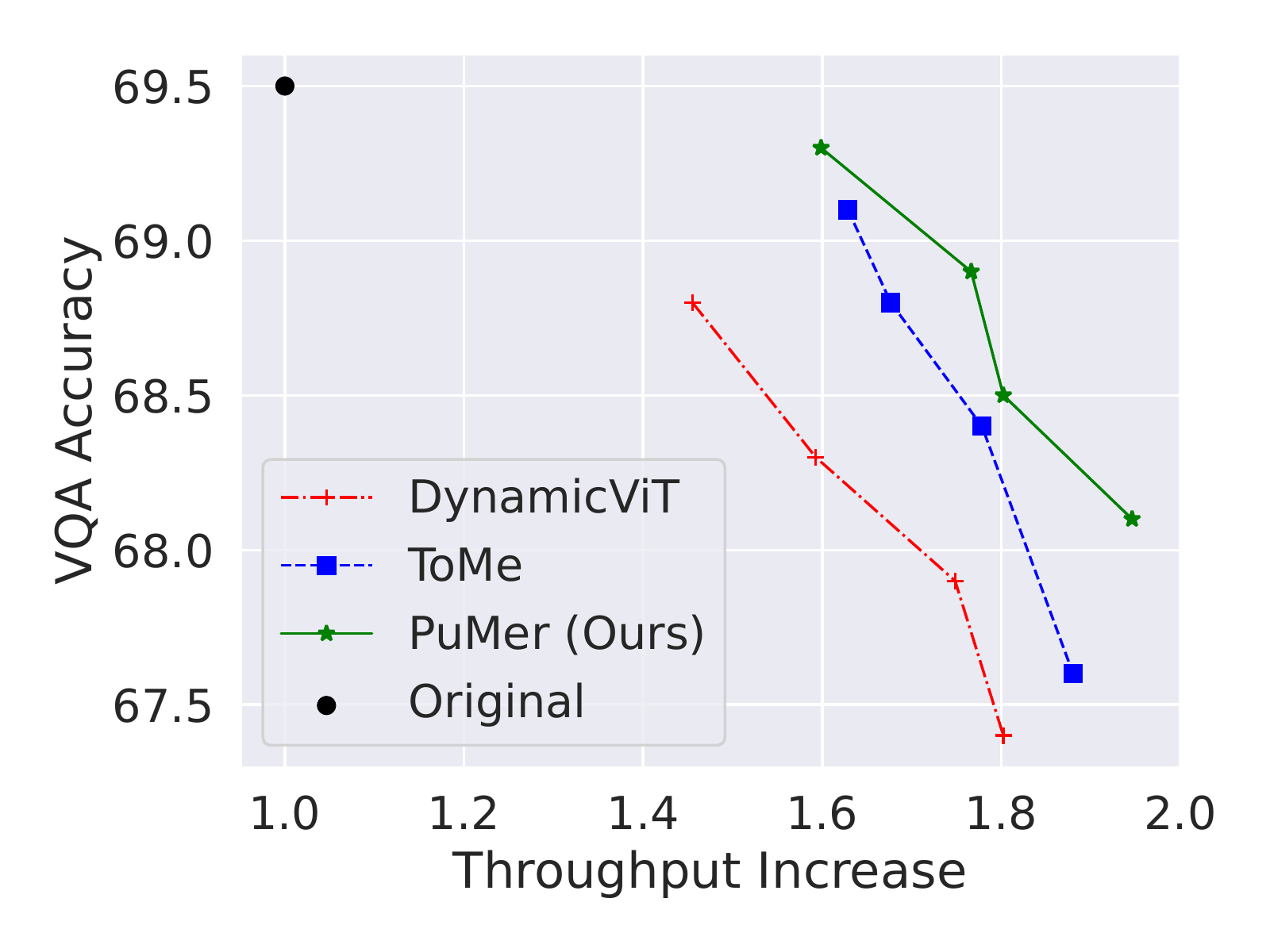}
		\caption{Comparing \sys with DynamicViT and ToMe for the ViLT model on the VQAv2 dataset. Setting different pruning and merging ratios for DynamicViT and ToMe gives different inference throughput and accuracy numbers. Right and top lines are better trade-offs.}
		\label{fig:baselines}
	\end{center}
\end{figure}

\paragraph{\sys is more accurate and faster than previous token reduction methods.} \cref{fig:baselines} presents the accuracy versus inference throughput increase trade-offs for \sys, DynamicViT and ToMe applied to the ViLT model on the VQAv2 dataset. Given a similar throughput increase (like 1.8x), \sys has the best accuracy compared to DynamicViT and ToMe. Similarly, for a given accuracy drop constraint (like $<$ 1\%), \sys provides a bigger throughput increase.

\paragraph{\sys provides larger efficiency gains over smaller resolution baselines.}

\begin{table}[t]
    \centering
    \setlength\tabcolsep{3pt}
    \small
\begin{tabular}{@{}lllcc@{}}
\toprule
\multirow{2}{*}{Model} & Image    & VQAv2  & Throughput  & Memory  \\ 
& Resolution & Accuracy  & Increase  & Reduction \\ \midrule
\multirow{4}{*}{Resolution} & 192x192            & 74.3 (-3.2)                     & 4.23x                             & 75\%                            \\
& 224x224            & 75.2 (-2.3)                     & 3.48x                             & 66\%                            \\ 
& 256x256            & 76.1 (-1.4)                     & 2.67x                             & 54\%                            \\
& 320x320            & 77.0 (-0.5)                     & 1.62x                             & 37\%                            \\ \midrule
\textbf{\sys} & 320x320   & 76.3 (-1.2)   & 2.86x    & 59\% \\ \midrule

\textbf{\sys} & 384x384   & 76.8 (-0.7)   & 1.82x    & 38\% \\ \midrule
Original        &  384x384  & 77.5          & 1x       & 0\% \\ \bottomrule
\end{tabular}
\caption{Performance and inference efficiency comparison between the smaller resolution baselines and \sys for the METER model on the VQAv2 test set.}
\label{table:baseline}
\end{table}

\cref{table:baseline} shows the results for the METER model on the VQAv2 dataset when comparing \sys with downsampling the input image to smaller resolutions. Using smaller resolution input images improves the inference throughput and reduces memory footprint but comes with larger accuracy drops. The closest resolution is 320x320 which is slightly more (0.2\%) accurate than \sys, but it has 20\% lower inference throughput.  
Meanwhile, \sys is orthogonal to downsampling strategies, and applying \sys to smaller images could provide additional efficiency gains; for input image resolution 320x320, \sys improves METER throughput by 1.76x with a 0.7\% accuracy drop\footnote{1.76=2.86/1.62, 0.7=77.0-76.3}  (see the 3rd row numbers in \cref{table:baseline}).

\subsection{Ablation Study}
\label{sec:ablation}
\paragraph{Effectiveness of \sys Components.}

\begin{table}[ht!]
    \centering
    \setlength\tabcolsep{3pt}
    \small
\begin{tabular}{@{}llc@{}}
\toprule
\multirow{2}{*}{Model}                           & VQA  & Throughput  \\ 
 &  Accuracy &  Increase \\ \midrule
ViLT                            & 69.5       & 1x  \\ \midrule
\sys-ViLT                       & 68.9 (-0.6) & 1.76x   \\
\quad w/o text-informed image pruning   & 69.2 (-0.3)  & 1.52x  \\
\quad w/o modality-aware merging           & 69.1 (-0.4) & 1.46x   \\
\quad w/o distillation               & 68.6 (-0.9) & 1.76x   \\
\bottomrule
\end{tabular}
    \caption{Ablation analysis for each component in \sys on the VQAv2 dataset for ViLT model.}
    \label{tab:ablation-arch}
\end{table}

\begin{table*}[ht!]
    \centering
    \setlength\tabcolsep{3pt}
    \small
\begin{tabular}{@{}llccclc@{}}
\toprule
Choice & Reduction Layers & Prune Ratio & Image Merge Ratio & Text Merge Ratio & VE Accuracy & Throughput Increase \\ \midrule
\multirow{4}{*}{ratios} & 2,5,8  & 0.1 & 0.3 & 0.2  & 75.8 (-0.2)  & 1.77x \\
                        & 2,5,8  & 0.3 & 0.3 & 0.2  & 74.7 (-1.3)  & 2.04x \\ 
                        & 2,5,8  & 0.1 & 0.3 & 0.5  & 74.9 (-1.1)  & 1.89x \\ 
                        & 2,5,8  & 0.1 & 0.5 & 0.2  & 73.8 (-2.1)  & 2.12x \\ \midrule
\multirow{3}{*}{\# of layers} & 2 & 0.1 & 0.3 & 0.2  & 75.9 (-0.15)  & 1.43x \\ 
                              & 2,4 & 0.1 & 0.3 & 0.2 & 75.8 (-0.2)  & 1.69x \\ 
                              & 2,4,6 & 0.1 & 0.3 & 0.2 & 75.7 (-0.3)  & 1.80x \\ \midrule
\multirow{2}{*}{locations} & 2,3,4 & 0.2 & 0.2& 0.2 & 74.2 (-1.8) & 2.03x \\
                           & 7,8,9 & 0.2 & 0.2 & 0.2 & 75.9 (-0.1) & 1.31x \\ \midrule
\sys (Ours)                & 2,4,6,8 & 0.1 & 0.3 & 0.2 & 75.6 (-0.4) & 2.01x \\\midrule
ViLT                       & - & -  &  - & - & 76.0 & 1.00x   \\\bottomrule
\end{tabular}
    \caption{Design choices analysis of prune and merge ratios, \# of reduction layers, and reduction locations for the ViLT model on SNLI-VE task.}
    \label{tab:ablation-pruning}
\end{table*}
To show how each component in \sys affects the VL task accuracy and model inference efficiency, we ablate the three components --- text-informed image pruning, modality-aware merging and distillation --- in \cref{tab:ablation-arch}. %
Applying text-informed image pruning or modality-aware merging individually has shown improvements in model inference throughput with smaller accuracy loss. But stacking the two techniques together provides bigger inference efficiency without losing much task performance. Without knowledge distillation, \sys is still accurate and fast and adding it further reduces the performance gap.

\paragraph{Token Reduction Design Choices.} 
\label{sec:prune}

Given a 12-layer VL cross-modal encoder like ViLT, many combinations of reduction locations and ratios achieve similar inference speedups. Reducing tokens at earlier layers with lower ratios has similar computation efficiency to pruning at later layers with higher ratios. For comparing the accuracy with different numbers of reduction layers, we control the inference throughput to be similar to \sys by selecting the pruning and merging ratios and locations. \cref{tab:ablation-pruning} shows cascaded reduction at 4 layers (2th, 4th, 6th, 8th) has higher accuracy and speedups.

The ratios row in \cref{tab:ablation-pruning} shows reducing (via pruning or merging) more tokens leads to a bigger throughput increase but has a significant ($>$1\%) accuracy drop while reducing fewer tokens is more accurate but causes lower throughput. As shown in the locations row, we find that reducing tokens in the earlier layers leads to bigger throughput but drops accuracy by 1.8\%, while reducing tokens in the later layers is slightly more accurate but provides fewer benefits in throughput. Overall, for ViLT on the SNLI-VE task, we choose a 4-layer cascaded token reduction strategy with a pruning ratio of 0.1 and merging ratio of 0.3 and 0.2 for image and text respectively, and scatter the reduction locations more evenly to balance accuracy and speed trade-offs.

\section{Conclusion}
\label{sec:conclusion}
Large vision language models have been effective at visual reasoning tasks due to their complex cross-modal interactions between the text and image tokens. These cross-modal interactions are computationally expensive because all image and text tokens are processed in many layers. We introduce a token reduction framework --- \sys that uses text-informed image pruning and modality-aware merging techniques to effectively reduce the image and text tokens inside cross-modal layers. \sys progressively removes the redundant image and text information and makes VL models run faster with minimal task performance drop. \sys is easy to train and speeds up both training and inference of vision and language models across diverse downstream visual reasoning tasks. %

\section*{Acknowledgements}
This research was supported partly by NSF IIS-2044660, an Allen Investigator Distinguished award.
We thank the anonymous reviewers and the members of the UW NLP group for their comments and feedback on this paper.

\section{Limitations}
\label{sec:limit}

Our method does not apply to VL models where the cross-modal encoder layers are relatively lightweight. For example, the vision encoder is much more computationally expensive than the cross-modal encoder for VL models like ALBEF~\cite{liAlignFuseVision2021} and X-VLM~\cite{zengMultiGrainedVisionLanguage2021}, therefore, the end to end inference speed improvement is marginal. Reducing the image tokens inside the vision encoder could further improve the model efficiency, we leave this exploration to future work.

\bibliography{ref}
\bibliographystyle{acl_natbib}
\clearpage

\appendix

\section{Appendix}
\label{sec:appendix}

\subsection{\sys Details}
\label{sec:impl}

\paragraph{Implementation.}

We use the Transformers~\cite{wolf-etal-2020-transformers} and Accelerate~\cite{HuggingfaceAccelerate2022} with DeepSpeed~\cite{DeepSpeed} library to implement the training tasks. We conduct training jobs on 4 Nvidia A100 GPUs. 
For both ViLT and METER model, we first follow the training hyperparameters in their original papers and finetune the pretrained model to obtain task-specific models. These models are used as baselines for measuring accuracy drop and also used as the teacher model for \sys distillation. For baseline VL models, we finetune both METER and ViLT models on the studied VL tasks for 10 epochs. For \sys, we finetune 20 epochs using early stopping with a penitence of 5 (the accuracy won't improve after 5 epochs).

We list all training hyperparameters in \cref{tab:hyper}.
\begin{table}[ht!]
    \centering
    \setlength\tabcolsep{2pt}
    \small
\begin{tabular}{@{}lcccc@{}}
\toprule
METER & Retrieval & VQAv2 & NLVR2   & SNLI-VE  \\\midrule                     
cross-modal lr   & 2.5e-5  & 2.5e-5 & 5e-5  & 1e-5   \\
classifier lr & 2.5e-5 & 2.5e-4 & 1e-4       & 2e-5  \\
batch size per gpu     & 32    & 32     & 16   & 32     \\
image size              & 384        & 384    & 288   & 384   \\
 patch size               & 16       & 16     & 16     & 16     \\ \bottomrule
\end{tabular}
\begin{tabular}{@{}lcccc@{}}
\toprule
ViLT & Retrieval & VQAv2 & NLVR2  & SNLI-VE   \\\midrule                     
cross-modal lr     & 1e-4 & 1e-4        & 1e-4 & 1e-4       \\
classifier lr & 1e-4 & 1e-3       & 1e-4 & 1e-3       \\
batch size per gpu         & 32     & 64        & 32  & 64            \\
image size                    & 384    & 384         & 384  & 384        \\
 patch size                    & 32     & 32          & 32   & 32         \\ \bottomrule
\end{tabular}
\caption{Hyperparameters for finetuning \sys and original VL models.}
\label{tab:hyper}
\end{table}

We list the default reduction layers and ratios for different VL tasks in \cref{tab:ratios}.

\begin{table}[ht!]
    \centering
    \setlength\tabcolsep{3pt}
    \small
\begin{tabular}{@{}lllll@{}}
\toprule
METER        & VQAv2   & NLVR2 & SNLI-VE & Retrieval \\ \midrule
Reduction Layers  & 0,2,4,6 & 2,4,6 & 0,2,4,6 & 2,4,6 \\
Prune Ratio       & 0.2     & 0.3   & 0.3     & 0.2  \\
Image Merge Ratio & 0.2     & 0.5   & 0.5     & 0.5 \\
Text Merge Ratio  & 0.2     & 0.2   & 0.2     & 0.2  \\ \bottomrule
\end{tabular}
\begin{tabular}{@{}lllll@{}}
\toprule
ViLT         & VQAv2 & NLVR2 & SNLI-VE & Retrieval \\ \midrule
Reduction Layers  & 2,5,8 & 2,5,8 & 2,4,6,8 & 2,5,8     \\
Prune Ratio       & 0.1   & 0.1   & 0.1     & 0.1       \\
Image Merge Ratio  & 0.3   & 0.3   & 0.3     & 0.3       \\
Text Merge Ratio   & 0.2   & 0.2   & 0.2     & 0.2       \\ \bottomrule
\end{tabular}
\caption{Reduction layers and ratios for \sys-METER and \sys-ViLT on the VL tasks.}
\label{tab:ratios}
\end{table}

\subsection{Model Inference FLOPs Comparison}

\label{sec:flops}

\begin{table}[ht!]
    \centering
    \setlength\tabcolsep{5pt}
    \small
\begin{tabular}{@{}llrrr@{}}
\toprule
Model                                      & Datasets & Original & \sys &  Speedup \\ \toprule
{\multirow{3}{*}{METER}} & VQAv2    & 92                                  & 64.7                                           & 1.42x                               \\
                       & SNLI-VE  & 92                                  & 59                                             & 1.56x                               \\
                       & NLVR2    & 184                                 & 131                                            & 1.40x                               \\ \midrule
\multirow{3}{*}{ViLT}                      & VQAv2    & 16                                  & 8.7                                            & 1.84x                               \\
                                           & SNLI-VE  & 16                                  & 7.7                                            & 2.08x                               \\
                                           & NLVR2    & 32                                  & 17.4                                           & 1.84x                               \\ \bottomrule
\end{tabular}
\caption{GFLOPs comparison between \sys and original VL models for METER and ViLT.}
\label{tab:flops}
\end{table}

We measure FLOPs of both \sys and the original model for METER and ViLT using the fvcore tool\footnote{\url{https://github.com/facebookresearch/fvcore/blob/main/docs/flop_count.md}}. The results are shown in \cref{tab:flops}.

\end{document}